\documentclass[lettersize,journal]{IEEEtran}
\usepackage{amsmath,amsfonts}
\usepackage{array}
\usepackage{textcomp}
\usepackage{stfloats}
\usepackage{url}
\usepackage{verbatim}
\usepackage{graphicx}
\usepackage{subfigure}
\usepackage{xcolor}
\usepackage{cite}
\usepackage{amsthm,amssymb}
\usepackage{booktabs} 
\usepackage{multirow} 
\usepackage[noend]{algpseudocode}
\usepackage{algorithmicx,algorithm}
\usepackage[colorlinks,
            linkcolor=blue,
            anchorcolor=blue,
            citecolor=blue]{hyperref}
\author{{Fanfu Xue, En Yu, Yantian Shen, Zhikun Hu, Hongjun Wang, Yang Yang, Xindi Wang and Jiande Sun}
\thanks{Fanfu Xue, Yantian Shen, Hongjun Wang and Yang Yang are with the School of Information Science and Engineering, Shandong University, Qingdao 266237, China (e-mail: \{fanfuxue, ytshen\}@mail.sdu.edu.cn, \{hjw, yyang\}@sdu.edu.cn)

En Yu is with the Faculty of Engineering and Information Technology, University of Technology Sydney, Sydney, NSW, Australia (e-mail: isenn.yu@gmail.com)

Zhikun Hu is with the School of Computer Science and Technology, Shandong University, Qingdao 266237, China (e-mail: huzhikun@zhiyang.com.cn)

Xindi Wang is with the School of Artificial Intelligence, Shandong University, Jinan 250100, China (e-mail: xindi.wang@sdu.edu.cn)

Jiande Sun is with the School of Computer Science and Artificial Intelligence, Shandong Normal University, Jinan 250358, China, and also with the Interdisciplinary Research Center of General Artificial Intelligence, Shandong Normal University, Jinan 250358, China (e-mail: jiandesun@hotmail.com).
}
\thanks{Corresponding authors: Xindi Wang and Jiande Sun}
}
\title{See-and-Reach: Precise Vision-Language Navigation for UAVs within the Field of View}

\begin{document}

\maketitle

\begin{abstract}
UAV Vision-Language Navigation (UAV-VLN) is typically formulated as a holistic search-and-reach problem, where long-range target discovery and final target approach are optimized and evaluated jointly. This formulation makes it difficult to assess a critical capability of aerial embodied agents, namely whether a UAV can accurately ground a visible target and translate vision-language evidence into precise 3D motion once the target enters its field of view. To address this limitation, we introduce UAV-VLN-FOV, a target-visible navigation task that isolates the see-and-reach stage and enables a more diagnostic evaluation of terminal reaching ability. We further propose 3DG-VLN, a vision-language waypoint prediction framework guided by dynamic 3D direction cues to enhance fine-grained visual grounding and spatial direction alignment for precise target reaching. Specifically, 3DG-VLN adaptively processes high-resolution front-view and downward-view observations to preserve fine-grained visual and geometric details for target grounding. It also updates the target-relative direction online during closed-loop navigation, allowing the agent to maintain spatial alignment with the target and reduce accumulated direction drift. To support this task, we construct a dedicated high-resolution benchmark which contains 2,717 trajectories with target-oriented high-level instructions, high-resolution front-view and downward-view egocentric observations, and continuous 3D waypoint annotations. Experiments show that 3DG-VLN outperforms competitive UAV-VLN baselines, achieving a 13.82\% improvement in success rate. Real-world trials further demonstrate the potential of 3DG-VLN for practical see-and-reach navigation. The source code and benchmark are available at https://github.com/xuefanfu/3DG-VLN.
\end{abstract}


\section{Introduction}

\begin{figure}[t!]
\centering
\includegraphics[width=\linewidth]{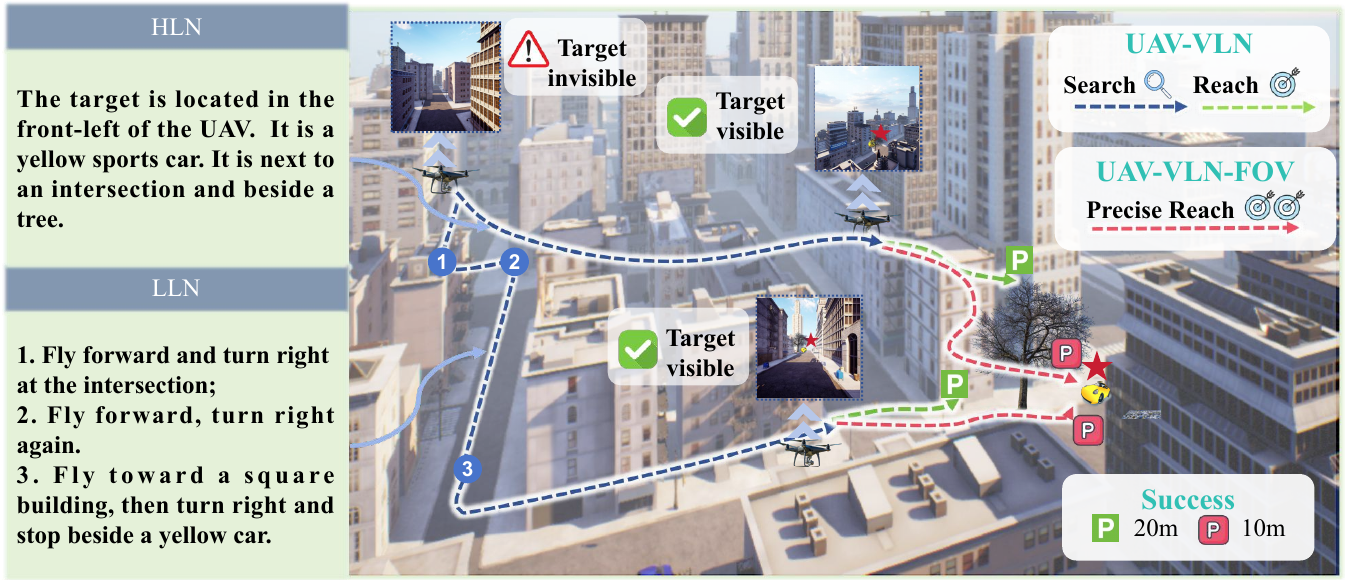}
\caption{UAV-VLN vs UAV-VLN-FOV. UAV-VLN emphasizes the holistic search-and-reach process, whereas UAV-VLN-FOV focuses on the see-and-reach stage.}
\label{different_task}
\end{figure}

UAV Vision-Language Navigation (UAV-VLN) requires an aerial agent to execute complex navigation tasks based on linguistic instructions and egocentric observations. Compared with ground-based VLN~\cite{wang2026expand,he2026fine,yu2025mossvln,tan2025source}, UAV-VLN presents unique challenges arising from the 3D topology of aerial spaces, which are inherently more expansive and less structured. Successful navigation in this domain necessitates the convergence of multimodal grounding, 3D spatial awareness, and dynamic motion planning under continuous control constraints. Despite recent exploratory efforts in instruction design, dataset construction and model frameworks~\cite{wang2024towards,ding2026history,zheng2026onfly}, current research often struggles to maintain the fine-grained spatial alignment required for complex real-world deployments, placing the field at a nascent stage where terminal reaching precision remains a largely unaddressed challenge.

Existing UAV-VLN studies~\cite{liu2023aerialvln,ning2026lookasidevln,wang2024towards} can be broadly categorized by instruction granularity into Low-Level Instruction Navigation (LLN) and High-Level Instruction Navigation (HLN). While LLN minimizes decision ambiguity through granular, sequential commands, such step-by-step guidance is often unrealistic in natural human-UAV interaction~\cite{zhang2026you}. Conversely, HLN abstracts navigation into semantic target descriptions, encompassing categories, visual attributes, and contextual relationships, rendering it more amenable to practical deployment. Despite utilizing diverse instruction formats, extant research predominantly conceptualizes UAV-VLN as a monolithic \textit{search-and-reach} process, as shown in Fig.~\ref{different_task}. We argue that this holistic formulation tends to obscure a fundamental divergence in navigational requirements, specifically the transition from \textit{semantic exploration} to \textit{geometric precision}. 

Specifically, we posit that the \textit{search stage} prioritizes high-level scene understanding and long-horizon discovery. In contrast, the \textit{reaching stage}, hereafter referred to as the \textit{see-and-reach} stage, necessitates an entirely distinct suite of capabilities including fine-grained visual localization and 3D trajectory control. The critical nature of this stage is exemplified in high-stakes missions such as post-disaster relief or emergency supply delivery. In these scenarios, once a target, such as a specific emergency supply station, is visually identified, the mission’s success hinges entirely on the agent’s ability to execute a precise approach for safe payload delivery. While current UAV-VLN benchmarks often adopt a permissive 20-meter success radius~\cite{liu2023aerialvln,wang2024towards,xiao2025uav}, such a coarse metric is inadequate for real-world missions where a 20-meter deviation could result in mission failure, such as dropping critical supplies into inaccessible terrain. Consequently, the prevailing holistic evaluation framework suffers from a lack of diagnostic granularity, as terminal precision failures are frequently masked by exploratory errors. By isolating the reaching process and advocating for a more stringent 10-meter success criterion, we seek to address a fundamental question in aerial embodied intelligence: \textbf{how an agent can effectively transform linguistic instruction and visual evidence into the precise 3D maneuvers required for reliably reaching the target vicinity.}

Motivated by these observations, we formalize \textbf{UAV-VLN-FOV}, a navigation task explicitly constrained by the agent’s \textit{Field of View} (FOV). This task concentrates on the \textit{see-and-reach} stage of aerial navigation, shifting the research emphasis from expansive exploration to precise target reaching once the objective becomes visually accessible, as shown in Fig.~\ref{different_task}. In this setting, the target is visible within the UAV's field of view at the initial position, necessitating the direct translation of linguistic instruction and egocentric visual evidence into executable 3D waypoint plans. To ensure a successful approach, the UAV must maintain robust spatial alignment despite radical scale variations, partial occlusions, and transient target disappearance from the field of view, while simultaneously executing reactive obstacle avoidance and continuous waypoint planning. To distinguish high-precision navigation from mere regional proximity, \textbf{UAV-VLN-FOV} mandates a stringent 10-meter success radius. By establishing this rigorous criterion, the task seeks to evaluate the agent’s terminal control fidelity in complex environments, ensuring that the navigational output meets the precision requirements of real-world deployment.


The setting reveals two critical deficiencies in extant methodologies. First, existing studies~\cite{wang2024towards,liu2023aerialvln} are primarily designed under long-range, search-oriented task settings, where low-resolution observations and visual architectures are often sufficient for scene-level exploration and coarse target localization. When applied to the precise reaching stage in UAV-VLN-FOV, they fail to reliably identify and approach targets. This is because low-resolution observations cannot precisely capture fine-grained geometric details, discriminate targets from visually similar distractors, or perceive nearby obstacles, all of which are essential for waypoint-level planning and terminal maneuvering. Second, existing direction-guided methods~\cite{wang2024towards,wu2025aeroduo} often rely on static initial cues. While these methods provide a coarse prior at the trajectory onset, they inevitably succumb to accumulative spatial drift as the UAV moves, leading to a severe misalignment between the linguistic instruction and the current egocentric perspective.

To address these challenges, we propose \textbf{3DG-VLN}, a vision-language waypoint prediction framework guided by dynamic 3D direction cues. Primarily, \textbf{3DG-VLN} processes high-resolution front-view and downward-view observations in a resolution-adaptive manner rather than through traditional down-sampling~\cite{wang2024towards,xu2026aerialvla}. This architecture enables the model to exploit fine-grained visual cues regarding target morphology and local spatial structures, thereby supporting more accurate visual grounding and 3D motion prediction. Additionally, to mitigate the risk of spatial drift, we introduce an online direction updating strategy that dynamically re-estimates the relative target direction from current observations. This mechanism ensures continuous spatial alignment and significantly improves reaching precision within the \textit{field of view}. To support training and evaluation under the UAV-VLN-FOV setting, we further construct a high-resolution benchmark driven by target-oriented high-level instructions. The dataset contains 2,717 trajectories and 31,878 image-instruction-waypoint tuples. It covers 14 realistic rural, town, and urban scenes, as well as 89 target categories derived from~\cite{wang2024towards}. Each trajectory consists of a target-oriented instruction, high-resolution front-view and downward-view observations, and corresponding 3D waypoint sequences. 
The main contributions of this work are summarized as follows:
\begin{itemize}
\item We formalize \textbf{UAV-VLN-FOV}, a novel task designed to evaluate the reaching-stage capability of aerial agents. By isolating the \textit{see-and-reach} stage and advocating for a stringent 10-meter success criterion, this task enables the systematic study of how agents achieve high-precision 3D navigation toward visually observable objectives.

\item We propose \textbf{3DG-VLN}, a vision-language waypoint prediction framework that leverages high-resolution dual-view observations for fine-grained target grounding and waypoint-level motion prediction. To address direction misalignment during flight, 3DG-VLN further updates the target direction online from current onboard observations and uses the updated cue to guide subsequent waypoint prediction.

\item We construct a high-resolution UAV-VLN-FOV benchmark with target-oriented high-level instructions, front-view and downward-view egocentric observations, and continuous 3D waypoint annotations, providing a dedicated basis for training and evaluating UAV agents in precise \textit{see-and-reach} navigation.

\item Extensive experiments show that 3DG-VLN achieves better target-reaching performance than existing UAV-VLN baselines under the UAV-VLN-FOV setting. Real-world experiments further demonstrate its practical applicability in \textit{see-and-reach} scenarios.
\end{itemize}

\section{Related Work}

\subsection{Aerial Navigation Datasets}

VLN has been extensively studied in ground-based embodied environments, where agents are required to follow natural language instructions by grounding them in visual observations and action histories~\cite{zhong2026run,wu2024vision,wang2026dynam3d,zhang2025flexvln,yu2026enhancing,zhang2025cosmo,xu2026vgas}. Inspired by these advances, recent efforts have extended VLN from ground robots to UAVs, giving rise to UAV-VLN. Compared with ground-based VLN, UAV-VLN introduces more challenging navigation conditions, including large-scale outdoor scenes, sparse visual landmarks, six-degree-of-freedom motion, and continuous control in 3D space.

AerialVLN~\cite{liu2023aerialvln} constructs a dedicated benchmark for language-guided aerial navigation, thereby establishing the foundation for this emerging field. Following this, AVDN~\cite{fan2023aerial} extends aerial navigation to a dialog-based setting, where the UAV can request additional guidance from a commander during navigation. OpenUAV~\cite{wang2024towards} further moves toward realistic UAV-VLN by building a simulation platform with multi-view observations and approximately 12k target-oriented trajectories, and introduces the UAV-Need-Help benchmark to study assistant-guided object search in complex aerial environments. AirSim360~\cite{ge2025airsim360} further enriches the perception setting by introducing panoramic observations for UAV navigation.

In addition, to reduce the gap between synthetic benchmarks and realistic urban aerial scenarios, CityNav~\cite{lee2024citynav} constructs a UAV-VLN dataset based on real urban-scale point clouds from SensatUrban~\cite{hu2022sensaturban}, providing realistic city-scale environments for aerial navigation. NavAgent~\cite{liu2024navagent} focuses on real urban street-view scenes and introduces a fine-grained landmark dataset designed to support landmark-centric UAV embodied navigation. More recently, AirNav~\cite{cai2026airnav} constructs a large-scale real-world UAV-VLN benchmark from real urban aerial data with natural and diverse instructions, addressing the limitations of prior datasets that mainly rely on virtual environments and templated language.

To better reflect the unique motion dynamics of UAVs, several datasets formulate UAV-VLN as continuous waypoint prediction rather than discrete action selection. OpenUAV~\cite{wang2024towards} and DuAl-VLN~\cite{wu2025aeroduo} emphasize continuous 6-DoF aerial motion and target-oriented navigation. In particular, DuAl-VLN introduces a dual-altitude collaborative setting, where a high-altitude UAV performs broad environmental reasoning while a low-altitude UAV conducts precise navigation and target grounding. Wang et al.~\cite{wang2025uav} further study short-distance language-conditioned UAV control in real-world scenarios, providing a benchmark for fine-grained trajectory execution.

Despite these advances, existing UAV-VLN datasets are mainly designed for either long-horizon target search, global path planning, dialog-guided exploration, or short-range specific actions. They do not explicitly isolate the critical \emph{see-and-reach} stage, where the target is already visible within the UAV's field of view and the agent must perform fine-grained visual grounding, local spatial reasoning, and precise waypoint-level motion planning. Therefore, they are not sufficient for evaluating whether a UAV can accurately approach a visible target under concise high-level instructions, which motivates our UAV-VLN-FOV benchmark.

\subsection{Aerial Navigation Agents}

Earlier UAV-VLN studies largely follow the ground-based VLN paradigm, using task-specific encoders and policy networks~\cite{xue2025acds,YU2026112075} to map visual observations and language instructions to navigation actions. However, the aerial domain requires stronger spatial reasoning, long-horizon planning, and continuous motion control. To address these challenges, recent methods increasingly exploit large language models (LLMs), vision-language models (VLMs), and multimodal foundation models to improve instruction understanding, visual grounding, and planning ability~\cite{yu2026generalized,liu2025vln,zhang2025multimodal,wu2025vla,jiang2026spatialfly,li2025skyvln,yuan2025seeing}.

One representative paradigm is to directly integrate general-purpose LLMs or VLMs into aerial navigation systems. GeoNav~\cite{xu2025geonav} introduces a geospatially aware multimodal chain-of-thought prompting mechanism. It projects instruction-related semantic masks into a top-down representation and converts spatial relations into metric-aware textual prompts, enabling LLMs to reason over landmark navigation, target search, and precise localization. VLFly~\cite{zhang2025grounded} 
adopts a modular framework for open-vocabulary UAV navigation. In this framework, an LLM-based instruction encoder reformulates high-level language into structured prompts, a VLM-based goal retriever associates these prompts with goal images, and a waypoint planner generates executable trajectories from monocular egocentric observations. CityNavAgent~\cite{zhang2025citynavagent} leverages LLMs to decompose complex urban navigation tasks into hierarchical planning subtasks, reducing long-horizon planning difficulty and predicting intermediate waypoints for downstream execution.

To further incorporate external spatial structures or semantic memory to improve robustness, NavAgent~\cite{liu2024navagent} combines multi-scale environmental information, including global topological maps, medium-scale panoramas, and local fine-grained landmarks. It further fine-tunes a GLIP-based landmark recognizer to linguisticize visual landmarks and uses a dynamically growing scene topology graph to support large-VLM-based UAV embodied navigation. AeroDuo~\cite{wu2025aeroduo} introduces a collaborative dual-UAV framework, in which a high-altitude UAV equipped with a multimodal LLM performs broad target reasoning, while a low-altitude UAV executes lightweight multi-stage navigation and target grounding. Such methods demonstrate the benefit of structured spatial priors for large-scale aerial navigation, but their focus remains on search and global decision-making rather than target-visible precise reaching.

Recent studies also explore task-specific adaptation of VLMs through fine-tuning or reinforcement learning. FlightGPT~\cite{cai2025flightgpt} proposes a two-stage training pipeline for UAV-VLN: supervised fine-tuning on high-quality demonstrations to improve initialization and structured reasoning, followed by group relative policy optimization with a composite reward that considers goal accuracy, reasoning quality, and format compliance. OpenVLN~\cite{lin2025openvln} further investigates data-efficient reinforcement learning for open-world aerial VLN, optimizing VLMs under limited training data and introducing a value-based long-horizon planner for trajectory synthesis. AerialVLA~\cite{xu2026aerialvla} moves toward end-to-end vision-language-action modeling by mapping raw visual observations and fuzzy linguistic instructions directly to continuous control signals, while another AerialVLA-style online-dialogue framework~\cite{chen2026aerialvla} studies proactive question asking and history-aware spatial-temporal fusion for UAV navigation with online human interaction.

Although these agents have significantly advanced UAV-VLN, most of them are optimized for holistic search-and-reach navigation, where the target must first be discovered in a large-scale environment before being approached. In contrast, UAV-VLN-FOV focuses on the final target-visible reaching stage, where the primary challenges shift from where to search to how to precisely ground, approach, and stop near the visible target. This setting requires high-resolution egocentric perception, local obstacle awareness, waypoint-level 3D motion planning, and fine-grained stopping behavior, which are not sufficiently evaluated by existing aerial navigation agents and benchmarks. Moreover, existing aerial VLN methods often differ substantially in their observation modalities and action interfaces: some operate on top-down geospatial maps, some rely on fixed action primitives or task-specific execution protocols. As a result, they are not well suited to the precise target-reaching requirement of UAV-VLN-FOV, which demands closed-loop execution with continuous waypoint-level 3D motion planning from egocentric observations.

\begin{figure*}
\centering
\includegraphics[width=\linewidth]{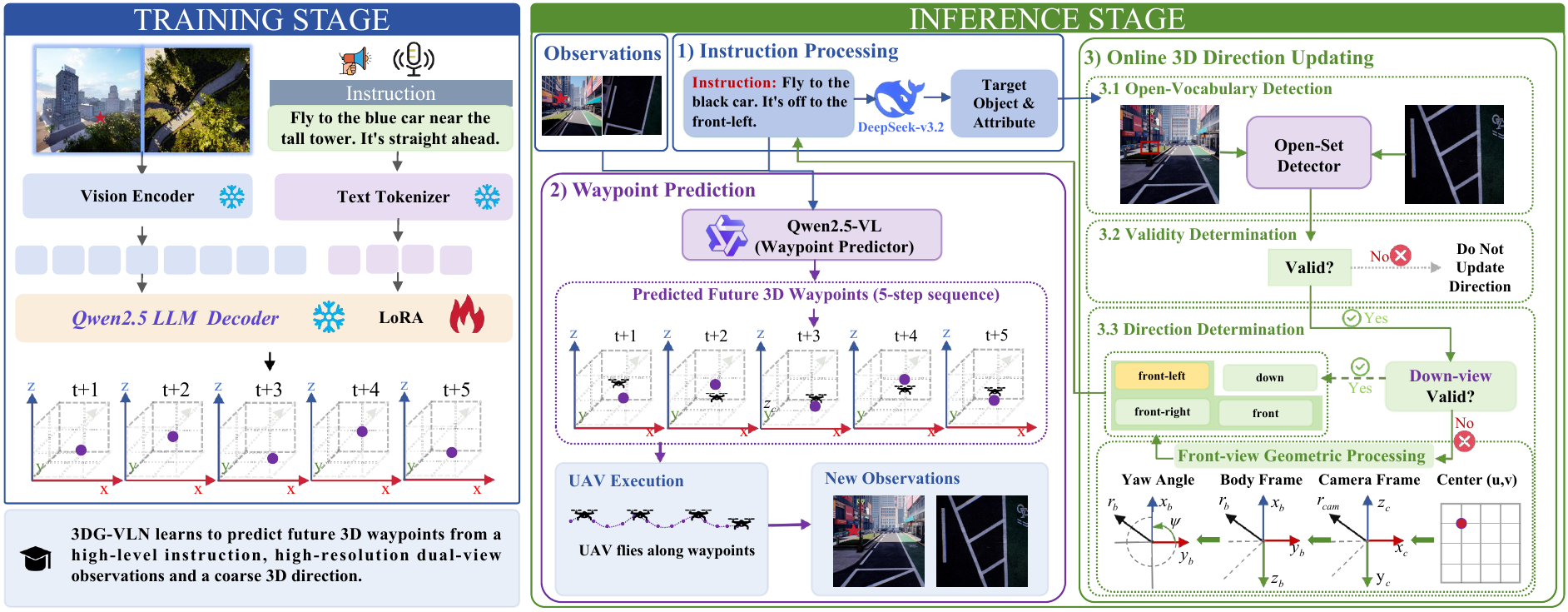}
\caption{Overall framework of 3DG-VLN. During training, we fine-tune Qwen2.5-VL using the constructed dataset to predict smooth waypoints based on instructions and observations. During inference, 3DG-VLN leverages the fine-tuned Qwen2.5-VL for waypoint prediction, DeepSeek V3.2 for extracting target objects and their intrinsic attributes, and an online 3D direction updating module for directional guidance.}
\label{network}
\end{figure*}

\section{UAV-VLN-FOV Task Formalization}
We define \textbf{UAV-VLN-FOV}, a UAV-VLN task within the \textit{field of view} for precise target reaching. Unlike beyond-visual-range exploration, \textbf{UAV-VLN-FOV} targets the final reaching stage where the objective is already visually accessible within the agent’s field of view. Formally, we formulate the \textit{see-and-reach} stage as a sequential 3D waypoint prediction problem for precise target reaching. Let $\mathcal{S} \subset \text{SE}(3)$ denote the state space of the UAV, where the pose at time $t$ is defined as $\mathbf{P}_t = (\mathbf{p}_t, \mathbf{q}_t)$, with $\mathbf{p}_t \in \mathbb{R}^3$ representing the Cartesian coordinates and $\mathbf{q}_t \in \mathbb{S}^3$ representing the orientation in the global frame. 

\textbf{Observation Space.} At each discrete timestep $t$, the agent receives a multimodal observation tuple $\mathbf{X}_t = \{ \mathcal{I}, \mathcal{V}_t, \mathcal{D}_t \}$. Concretely, $\mathcal{I}$ is the linguistic instruction, providing high-level semantic constraints. $\mathcal{V}_t = \{ \mathbf{R}^{front}_t, \mathbf{R}^{down}_t \}$ represents the high-resolution dual-view egocentric observations, where $\mathbf{R} \in \mathbb{R}^{H \times W \times 3}$. Distinct from conventional low-resolution exploration, our formulation preserves the \textit{perceptual details} of $\mathcal{V}_t$ needed to localize visually accessible targets, distinguish subtle distractors, and reason about nearby obstacles for waypoint-level planning. $\mathcal{D}_t \in \{ \textit{front, front-left, front-right, down} \}$ is the coarse 3D direction cue, which provides a semantically explicit spatial prior to maintain alignment between the agent and the target.

\textbf{Action Space and Policy Flow.} The agent's action is defined as the prediction of a short-horizon trajectory $\mathcal{W}_t = \{ \mathbf{w}_{t+k} \}_{k=1}^K$ on the waypoint manifold. To ensure translation and rotation invariance, each waypoint $\mathbf{w}_{t+k} = (\Delta x_{t+k}, \Delta y_{t+k}, \Delta z_{t+k})$ is expressed as a relative displacement in the UAV-centric local coordinate frame, where $\Delta x_{t+k}$, $\Delta y_{t+k}$, and $\Delta z_{t+k}$ denote forward, rightward, and downward displacements, respectively. The navigation policy $\pi$ executes a deterministic mapping from the observation history to the action space:
\begin{equation}
    \mathcal{W}_t = \pi( \mathcal{I}, \mathcal{V}_{\le t}, \mathcal{D}_t ; \theta )
\end{equation}
where $\theta$ denotes the learnable parameters. This policy flow requires the model to implicitly resolve the coordinate transformation between semantic visual evidence and physical actuation, mapping \textit{what is seen} into \textit{where to reach}.

\textbf{Objective and Success Criterion.} Given a ground-truth target position $\mathbf{p}_{tar}$, the navigational episode is successful if the terminal pose satisfies a stringent proximity constraint. The success indicator $S$ is defined as:
\begin{equation}
    S = \mathbb{I} \left( \| \mathbf{p}_T - \mathbf{p}_{tar} \|_2 \le \delta \right)
\end{equation}
where $\mathbf{p}_T$ is the terminal position of the UAV, and $\delta = 10$ meters represents the precision threshold necessitated by high-stakes real-world deployments (e.g., post-disaster relief or emergency supply delivery), which is stricter than the commonly used 20-meter threshold in previous UAV-VLN settings~\cite{liu2023aerialvln,wang2024towards,xiao2025uav}. The objective of \textbf{UAV-VLN-FOV} is to learn a policy $\pi$ that maximizes the expected success rate $\mathbb{E}_{\tau \sim \pi}[S(\tau)]$ within the \textit{field of view}, where the agent must handle large target-scale variations, viewpoint changes, and temporary loss of target visibility during precise reaching.

\section{Proposed Method}

\subsection{Overview}

As illustrated in Fig.~\ref{network}, we propose \textbf{3DG-VLN}, a vision-language waypoint prediction framework for the \textit{see-and-reach} stage of UAV-VLN. Given a language instruction, high-resolution visual observations and a coarse 3D direction cue, 3DG-VLN predicts continuous 3D waypoints in the UAV body frame. Our design focuses on two key aspects that are crucial for precise \textit{see-and-reach} navigation: \textbf{fine-grained visual grounding} and \textbf{spatial direction alignment}.

To achieve fine-grained visual grounding, 3DG-VLN employs Qwen2.5-VL to adaptively process high-resolution egocentric observations instead of aggressively downsampled inputs. This is important because, in the \textit{see-and-reach} stage, the target may be visible but still occupy a small image region or be identifiable only through fine-grained appearance and contextual details. By constructing multimodal instruction-response pairs with a language instruction, high-resolution visual observations, a coarse 3D direction cue, and waypoint responses, the model learns to associate the specified target with detailed visual evidence and translate such grounding into continuous 3D waypoint predictions, as detailed in Sec.~\ref{Training}.

To achieve spatial direction alignment, 3DG-VLN dynamically updates the coarse 3D direction cue during inference, aligning the model’s guidance with the UAV’s current egocentric observations, as detailed in Sec.~\ref{Inference}. Instead of treating the initial direction as a static prior, 3DG-VLN dynamically re-estimates the target's relative direction from the UAV's current observations and updates the direction cue online. This mechanism continuously aligns the model's spatial guidance with the current egocentric viewpoint, reducing accumulated direction drift and enabling more accurate target reaching.

\begin{figure}[]
\centering
\includegraphics[width=0.6\linewidth]{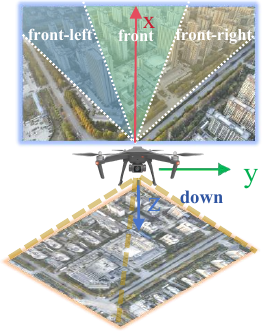}
\caption{3D direction schematic of 3DG-VLN. The front-left, front, front-right, and down bins describe the target’s relative position in the UAV-centric body frame, where the x, y, and z axes denote front, right, and down, respectively.}
\label{3D_direction}
\end{figure}

\subsection{3D Direction Construction}
\label{Direction}
To help the agent quickly anchor the target area under a large aerial field of view, we introduce a coarse 3D directional guidance system, as illustrated in Fig.~\ref{3D_direction}. Instead of using precise target angles or metric coordinates, the proposed representation discretizes the target direction into human-friendly direction cues, including \textit{front-left}, \textit{front}, \textit{front-right}, and \textit{down}. The coarse 3D direction serves as a unified directional interface throughout both training and inference. During training, it provides an explicit spatial condition that helps the model learn direction-aware waypoint generation. During inference, the same direction representation is used to maintain spatial consistency during closed-loop navigation in high-fidelity simulation and real-world environments, providing continuous alignment between the UAV's egocentric perception and the target. 

As shown in Fig.~\ref{3D_direction}, we define a UAV-centric body frame with \(x\), \(y\), and \(z\) axes corresponding to front, right, and down, respectively. Given the UAV pose (\(\mathbf{p}_{t}\),\(\mathbf{q}_{t}\)) and target position \(\mathbf{p}_{tar}\), the target displacement is first transformed from the global frame into the UAV body frame:
\begin{equation}
\mathbf{d}_{t} =
\mathbf{R}(\mathbf{q}_{t})^\top
\left(\mathbf{p}_{tar} - \mathbf{p}_{t} \right),
\label{eq:body_displacement}
\end{equation}
where \(\mathbf{R}(\mathbf{q}_{t})\) is the rotation matrix corresponding to the UAV orientation, mapping vectors from the UAV body frame to the global frame. Therefore, \(\mathbf{R}(\mathbf{q}_{t})^\top\) maps the target displacement from the global frame to the UAV body frame.

Let \(\mathbf{d}_{t}=({d}_{x},{d}_{y},{d}_{z})\), and let \(\|\mathbf{d}_{t}\|\) denote its Euclidean norm. The horizontal yaw angle of the target relative to the UAV heading is computed as
\begin{equation}
\alpha_{t} = \arctan2({d}_{y}, {d}_{x}).
\label{eq:target_yaw}
\end{equation}

The coarse 3D direction label \(\mathcal{D}_{t}\) is assigned by
\begin{equation}
\mathcal{D}_{t} =
\begin{cases}
\textit{down}, 
& \text{if } \dfrac{{d}_{z}}{\|\mathbf{d}_{t}\|} > \cos\phi, \\[6pt]
\textit{front-left}, 
& \text{else if } -\phi \leq \alpha_{t} < -\gamma_{\mathrm{front}}, \\[4pt]
\textit{front}, 
& \text{else if } -\gamma_{\mathrm{front}} \leq \alpha_{t}  \leq \gamma_{\mathrm{front}}, \\[4pt]
\textit{front-right}, 
& \text{else if } \gamma_{\mathrm{front}} < \alpha_{t}  \leq \phi,
\end{cases}
\label{eq:direction_rule}
\end{equation}
where $\phi$ denotes the outer boundary angle threshold of the coarse directional regions, \(\alpha_{t}\) represents the horizontal azimuth of the target on the UAV body-frame \(x\)-\(y\) plane at the timestep \({t}\), and \({d}_{z}/\|\mathbf{d}_{t}\|\) measures whether the target direction is dominated by the downward viewing component. The threshold \(\gamma_{\mathrm{front}}\) separates the central front region from the left and right regions.

This coarse 3D direction representation is not intended to provide precise geometric control. Instead, it offers a compact and human-interpretable spatial cue that guides the model toward the target region and provides a consistent direction space for subsequent waypoint prediction and online direction updating.

\subsection{Training Stage}  \label{Training}

The goal of the training stage is to adapt a general-purpose vision-language model into a geometry-aware waypoint prediction policy for UAV-VLN-FOV. 3DG-VLN is trained to map concise high-level instructions, \textbf{high-resolution egocentric observations}, and coarse 3D directional priors into sequences of 3D waypoints. Unlike approaches that downsample images to fit a low-resolution model~\cite{wang2024towards,xu2026aerialvla}, 3DG-VLN leverages a model capable of \textbf{adaptive high-resolution processing}, preserving fine-grained visual details critical for accurately perceiving targets and subtle environmental cues in UAV-VLN-FOV scenarios.

To achieve this, we employ Low-Rank Adaptation (LoRA)~\cite{hu2022lora} to efficiently fine-tune a pre-trained vision-language model Qwen2.5-VL~\cite{bai2025qwen2} for waypoint prediction. This approach allows the model to fully exploit high-resolution visual inputs without sacrificing pretrained multimodal reasoning abilities, ensuring that the extracted fine-grained visual features effectively guide the UAV's body-centric motion planning.

\paragraph{Multimodal Training Formulation}
For each training trajectory, the target-oriented instruction \(\mathcal{I}\) is kept fixed across sampled timesteps. The coarse 3D direction cue is constructed from the initial UAV pose and the target position according to Sec.~\ref{Direction}, and is used as a trajectory-level spatial prior during training. At timestep \(t\), the model takes the multimodal observation \(\mathbf{X}_t=\{\mathcal{I}, \mathcal{V}_t, \mathcal{D}_0\}\) as input, where \(\mathcal{V}_t=\{\mathbf{R}^{front}_t,\mathbf{R}^{down}_t\}\) denotes the high-resolution front-view and downward-view observations, and \(\mathcal{D}_0\) denotes the initial coarse 3D direction cue.

The target output is a sequence of $K$ future UAV-centric waypoints:
\begin{equation}
\mathcal{W}_t =
\left\{
\mathbf{w}_{t+k}
\right\}_{k=1}^{K}.
\end{equation}

Thus, each training sample is formulated as:
\begin{equation}
\left(
\mathbf{X}_t,\,
\mathcal{W}_t
\right)
=
\left(
\left\{
\mathcal{I},\,
\mathcal{V}_t,\,
\mathcal{D}_0
\right\},
\left\{
\mathbf{w}_{t+k}
\right\}_{k=1}^{K}
\right).
\end{equation}

This formulation encourages the model to ground the language-specified target in high-resolution visual observations, use the direction cue as a coarse spatial prior, and generate future motion in a form directly executable by the UAV controller.

\paragraph{Egocentric Waypoint Supervision}
To ensure translation and rotation invariance, we represent each future waypoint in the UAV-centric body coordinate system.

For a future timestep $t+k$, the displacement from the current UAV position to the future position in the global frame is computed as
\begin{equation}
\Delta \mathbf{p}^{\mathrm{world}}_{t+k}
=
\mathbf{p}_{t+k}
-
\mathbf{p}_{t}.
\end{equation}

Next, we transform this displacement into the UAV body frame:
\begin{equation}
\mathbf{w}_{t+k}
=
\mathbf{R}(\mathbf{q}_t)^\top
\Delta \mathbf{p}^{\mathrm{world}}_{t+k}.
\end{equation}

Finally, the resulting waypoint $\mathbf{w}_{t+k} = (\Delta x_{t+k}, \Delta y_{t+k}, \Delta z_{t+k})$ represents the relative 3D displacement from the UAV’s current pose to a future trajectory point. By supervising the model with the sequence $\{\mathbf{w}_{t+k}\}_{k=1}^{K}$, 3DG-VLN learns to generate short-horizon, body-frame waypoint predictions that align with egocentric observations and are suitable for closed-loop aerial navigation. 

\paragraph{Training Objective}
We follow the standard instruction-tuning paradigm and serialize the waypoint sequence into a structured assistant response. Given the multimodal input
 \(\mathbf{X}_t\), the model is trained to autoregressively generate the target waypoint response
$\mathcal{S}_t=(s_{t,1},\dots,s_{t,L_t})$.
The supervised fine-tuning loss is defined as
\begin{equation}
\mathcal{L}_{\mathrm{SFT}}
=
-
\frac{1}{N}
\sum_{t=1}^{N}
\frac{1}{L_t}
\sum_{\ell=1}^{L_t}
\log p_{\theta}
\left(
s_{t,\ell}
\mid
\mathbf{X}_t, s_{t,<\ell}
\right),
\end{equation}
where $p_{\theta}$ denotes the LoRA-adapted Qwen2.5-VL model, and $s_{t,<\ell}$ denotes the preceding response tokens. 


\subsection{Inference Stage} \label{Inference}

During inference, 3DG-VLN performs closed-loop \textit{see-and-reach} navigation, built upon the fine-tuned Qwen2.5-VL waypoint predictor. As illustrated in Fig.~\ref{network}, the inference framework integrates an instruction parser based on DeepSeek-V3.2, the fine-tuned waypoint predictor, and the online direction updating module. The instruction parser extracts the target object and its visual attributes, providing a semantic foundation for the direction updating module. At each timestep \(t\), the \textbf{online 3D direction updating module} dynamically estimates the coarse 3D direction from onboard high-resolution front-view and downward-view observations. The updated direction cue is incorporated into the multimodal observation
\(\mathbf{X}_t = \{\mathcal{I}, \mathcal{V}_t, \mathcal{D}_t\}\)
to guide the prediction of short-horizon waypoints. By iteratively updating the direction cue, the UAV continuously corrects for spatial drift, ensuring that subsequent waypoint predictions remain aligned with the target and improving overall target-reaching performance.

\paragraph{Closed-loop Waypoint Prediction}
At each timestep \(t\), the model receives the multimodal observation \(\mathbf{X}_t=\{\mathcal{I}, \mathcal{V}_t, \mathcal{D}_t\}\) with the coarse 3D direction cue \(\mathcal{D}_t\) being updated dynamically as the UAV navigates. The initial direction \(\mathcal{D}_0\) is provided at the start of each navigation episode. The fine-tuned Qwen2.5-VL predicts a sequence of $K$ future waypoints in the UAV body frame:
\begin{equation}
\left\{
\mathbf{w}_{t+k}
\right\}_{k=1}^{K}
=
p_{\theta}
\left(
\mathbf{X}_t
\right).
\end{equation}

To execute these predictions, each egocentric waypoint is transformed back to the global coordinate frame by reversing the waypoint construction in Sec.~\ref{Training}:
\begin{equation}
{\mathbf{p}}_{t+k}
=
\mathbf{p}_t
+
\mathbf{R}(\mathbf{q}_t)\mathbf{w}_{t+k},
\quad k=1,\dots,K.
\end{equation}

The UAV follows the resulting global waypoint sequence and then captures new front-view and downward-view observations at the updated timestep.

\paragraph{Online 3D Direction Updating}
To ensure accurate spatial guidance throughout the trajectory, the online direction updating module continuously processes front-view and downward-view images to detect the target. When a valid detection is obtained, the 3D direction is generated according to the downward-view priority principle. For a valid detection from the front-view camera, the module estimates the target’s relative 3D orientation using camera geometry. The center pixel of the bounding box is first converted into a normalized ray in the camera frame, which is then transformed into the UAV body frame to compute the horizontal yaw angle. This angle is discretized into a semantic direction category and incorporated into the instruction to guide the prediction of subsequent waypoints.

Specifically, given the target object and attributes extracted from the instruction, we apply Grounding DINO~\cite{liu2023grounding} to the front-view and downward-view images, producing detection sets $\mathcal{B}^{\mathrm{front}}$ and $\mathcal{B}^{\mathrm{down}}$.

Since multiple detections in the same view may introduce target ambiguity, we only regard a view as valid when it contains exactly one detection:
\begin{equation}
\tilde{\mathcal{B}}^{v} =
\begin{cases}
\mathcal{B}^{v}, & \text{if } |\mathcal{B}^{v}|=1, \\
\emptyset, & \text{otherwise},
\end{cases}
\quad
v\in\{\mathrm{front},\mathrm{down}\}.
\end{equation}
The valid observation for direction updating is selected as
\begin{equation}
(\mathcal{B}^{\mathrm{valid}}, v^{\mathrm{valid}})=
\begin{cases}
(\tilde{\mathcal{B}}^{\mathrm{down}}, \mathrm{down}),
& \text{if } \tilde{\mathcal{B}}^{\mathrm{down}}\neq \emptyset, \\[2pt]
(\tilde{\mathcal{B}}^{\mathrm{front}}, \mathrm{front}),
& \text{else if } \tilde{\mathcal{B}}^{\mathrm{front}}\neq \emptyset, \\[2pt]
(\emptyset, \mathrm{none}),
& \text{otherwise}.
\end{cases}
\end{equation}

This rule gives priority to the downward view because a valid detection from the downward-view camera indicates that the target is already located below the UAV, where vertical alignment becomes the dominant navigation cue.

The updated direction cue $\mathcal{D}_t$ is then determined by
\begin{equation}
\mathcal{D}_t =
\begin{cases}
\mathcal{D}_{t-1},
& \text{if } v^{\mathrm{valid}}=\mathrm{none}, \\[2pt]
\textit{down},
& \text{else if } v^{\mathrm{valid}}=\mathrm{down}, \\[2pt]
g_{\mathrm{front}}(\mathcal{B}^{\mathrm{valid}}),
& \text{otherwise  } v^{\mathrm{valid}}=\mathrm{front},
\end{cases}
\end{equation}
where $\mathcal{D}_{t-1}$ is the previous direction cue, and $g_{\mathrm{front}}(\cdot)$ maps a valid front-view detection to a horizontal direction category.

For a valid front-view detection box, let $(u,v)$ denote its center pixel. Given image resolution $W$, $H$, field of view $\mathrm{FOV}$, principal point $(c_x,c_y)=(W/2,H/2)$, and focal length $f =\frac{W}{2\tan(\mathrm{FOV}/2)}$, we first compute the normalized ray in the camera frame:
\begin{equation}
\mathbf{r}_{\mathrm{cam}}'
=
\left(
\frac{u-c_x}{f},
\frac{v-c_y}{f},
1
\right),
\quad
\mathbf{r}_{\mathrm{cam}}
=
\frac{\mathbf{r}_{\mathrm{cam}}'}{\|\mathbf{r}_{\mathrm{cam}}'\|}.
\end{equation}

The ray is then transformed into the UAV body frame:
\begin{equation}
\mathbf{r}_{\mathrm{body}}
=
\begin{bmatrix}
0 & 0 & 1 \\
1 & 0 & 0 \\
0 & 1 & 0
\end{bmatrix}
\mathbf{r}_{\mathrm{cam}}.
\end{equation}

Let $\mathbf{r}_{\mathrm{body}}=(r_x,r_y,r_z)$. The horizontal yaw angle of the detected target is
\begin{equation}
\psi =
\arctan2(r_y,r_x).
\end{equation}

Finally, the front-view direction is discretized as
\begin{equation}
g_{\mathrm{front}}(\mathcal{B}^{\mathrm{valid}})
=
\begin{cases}
\textit{front-left},
& \text{if } -\phi \leq \psi < -\gamma_{\mathrm{front}}, \\[2pt]
\textit{front},
& \text{if } -\gamma_{\mathrm{front}} \leq \psi \leq \gamma_{\mathrm{front}}, \\[2pt]
\textit{front-right},
& \text{if } \gamma_{\mathrm{front}} < \psi \leq \phi .
\end{cases}
\end{equation}

The updated direction cue is inserted into the multimodal observation \(\mathbf{X}_{t}=\{\mathcal{I}, \mathcal{V}_{t}, \mathcal{D}_{t}\}\), providing a precise spatially aligned prior for the model to generate subsequent waypoint predictions. Through this iterative process, 3DG-VLN continuously refines its spatial guidance from onboard observations and improves target-reaching precision.

\begin{figure}[]
\centering
\subfigure[Data split distribution]{
    \includegraphics[width=0.46\linewidth]{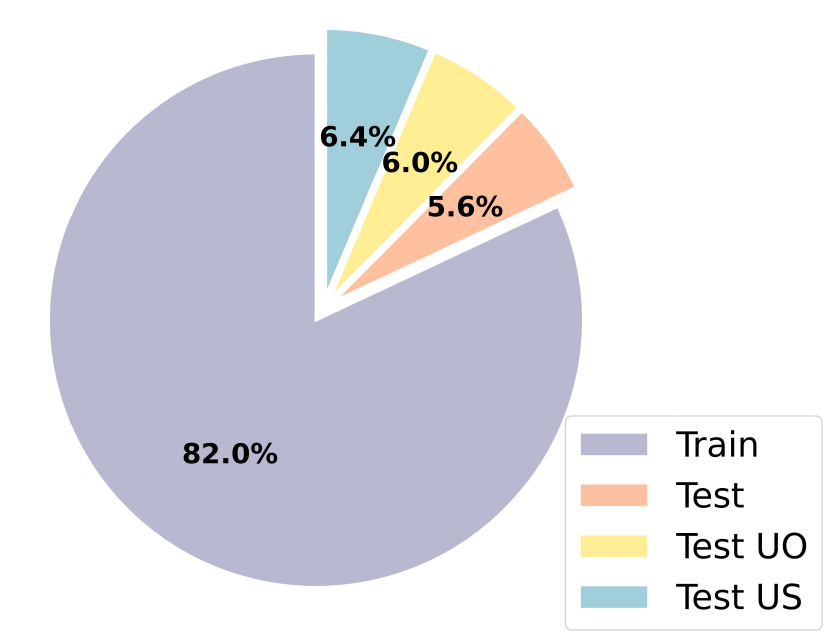}
    }
\subfigure[Trajectory length]{
    \includegraphics[width=0.46\linewidth]{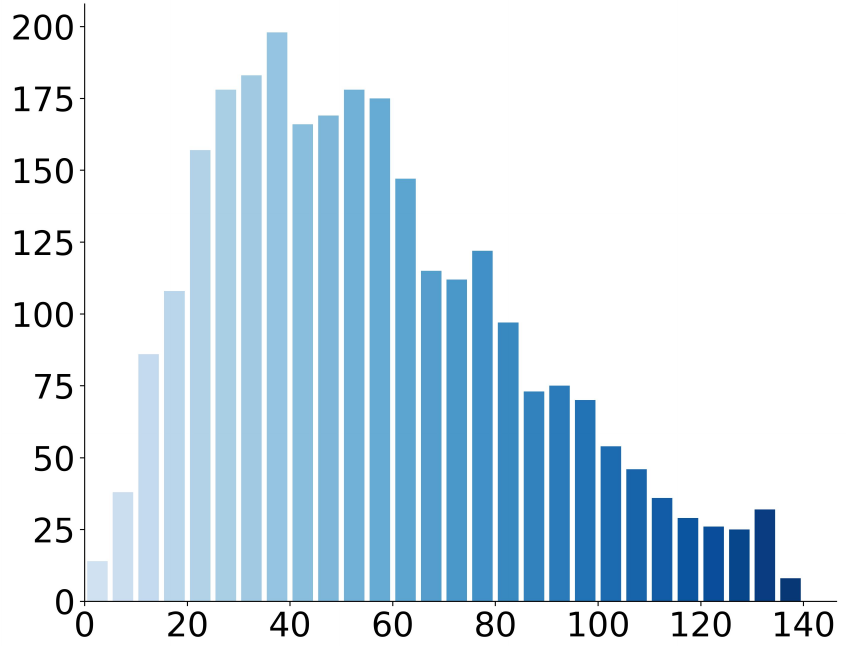}
    }
\caption{Dataset statistical analysis.}
\label{fig:direction_partition}
\end{figure}

\section{UAV-VLN-FOV Benchmark}
To facilitate the study of UAV-VLN-FOV, we construct a high-resolution benchmark driven by concise high-level instructions tailored for precise \textit{see-and-reach} navigation. The benchmark focuses on target-visible navigation, requiring agents to ground concise high-level instructions in high-resolution egocentric observations and produce waypoint-level plans for precise target reaching. This design complements existing UAV-VLN datasets centered on long-range search, offering a more diagnostic testbed for fine-grained grounding and local 3D planning.

\subsection{Dataset Construction}
We develop an asynchronous high-resolution imagery and continuous waypoint collection platform controlled via the AirSim API, leveraging the Unreal Engine environment, data assets and manually collected continuous waypoints provided in~\cite{wang2024towards}. 

Specifically, to maintain the diversity of target positions and prevent the agent from overfitting to specific target locations, we analyze the UAV-Need-Help dataset~\cite{wang2024towards} and count the number of trajectories corresponding to targets at each position in every scene. A maximum of two trajectories is selected from those corresponding to each position to construct a trajectory repository. Second, we develop the asynchronous high-resolution data collection platform based on the path-following asynchronous API in AirSim. Using this platform, we approximate the manually flown trajectories with the discontinuous waypoints from the UAV-Need-Help dataset, collect high-resolution images from the UAV’s forward and downward cameras at fixed time intervals, and sample the continuous flight trajectories simultaneously. Note that trajectories involving collisions during flight are discarded. Third, to build a high-quality UAV-VLN-FOV dataset, we manually screen the collected images to obtain trajectory sequences where the target appears within the filed of view. Furthermore, we leverage DeepSeek V3.2~\cite{liu2025deepseek} to convert the original target descriptions from the UAV-Need-Help dataset into application-friendly concise high-level instructions. Finally, we construct a high-resolution navigation dataset driven by concise high-level instructions, comprising 2,717 trajectories and 31,878 image-instruction-waypoint tuples.

\subsection{Dataset Analysis}
This dataset covers 14 different scenarios across three spatial categories: rural areas, towns, and cities. Significant differences exist in terrain, building density, and visual features among these scenarios, providing abundant data support for the UAV-VLN-FOV task. The dataset includes all 89 targets from~\cite{wang2024towards}. Following the division methods~\cite{wang2024towards,liu2023aerialvln}, we split the 2,717 generated trajectories into four subsets: the training set, test set (Test), unseen-object test set (Test UO) and unseen-scene test set (Test US). Specifically, the training set comprises 2,228 trajectories, while the Test, Test UO, and Test US sets contain 152, 164, and 173 trajectories, respectively, as shown in Fig.~\ref{fig:direction_partition} (a). Although this task is confined to the field of view, the overall trajectory length remains considerable due to the vast and unstructured outdoor space, with the maximum trajectory length reaching 138 meters, as shown in Fig.~\ref{fig:direction_partition} (b). In addition, the instructions in this dataset feature high practicality with an average length of merely 12 words, which is much shorter than that of the instructions in~\cite{wang2024towards,liu2023aerialvln} and better suited to real-world application scenarios.

\section{Experiments}
To comprehensively evaluate 3DG-VLN, we conduct extensive experiments on the UAV-VLN-FOV benchmark and in real-world environments, focusing on three key questions: (1) How does 3DG-VLN compare with SOTA UAV-VLN baselines? (2) Are the proposed components, including high-resolution observation processing, directional guidance, and online 3D direction updating, essential and effective for precise target reaching? (3) Can 3DG-VLN generalize from simulation to real-world scenarios? We answer these questions through comparisons with SOTA methods in Sec.~\ref{comparisons}, ablation studies in Sec.~\ref{ablation}, and real-world experiments in Sec.~\ref{real_world}.

\subsection{Configurations}

\noindent
\textbf{Baseline Methods.} 
We compare 3DG-VLN with four baselines: \textit{Random}, \textit{Fixed}, \textit{Qwen2.5-VL-7B}~\cite{bai2025qwen2} and \textit{TravelUAV}~\cite{wang2024towards}. \textit{Random} and \textit{Fixed} are used to provide lower-bound references, helping to quantify the navigable spatial range and the inherent difficulty of the \textit{see-and-reach} stage without intelligent guidance. \textit{Qwen2.5-VL-7B} is used as the base model of 3DG-VLN to examine the effect of task-specific adaptation for UAV-VLN-FOV waypoint prediction. \textit{TravelUAV} is adopted as the primary comparative baseline because it is the closest closed-loop waypoint-based UAV-VLN method to our setting. Unlike many existing aerial VLN agents that rely on different observation modalities, action spaces, or execution protocols, TravelUAV generates executable waypoint-level UAV motions under closed-loop navigation, enabling a fair comparison with 3DG-VLN in \textit{see-and-reach} stage. 

Specifically, \textit{1) Random} generates random 3D waypoints within the range of 2-5 meters based on the UAV's position at the current timestep $t$. \textit{2) Fixed:} requires that the UAV randomly selects actions from a fixed set, including straight ahead, shift left, shift right, turn left, turn right, ascend, descend, and stop. Among these, the distances for straight ahead, shift left, and shift right are set to 5 meters, the distances for ascend and descend are 2 meters, and the turn left and turn right angles are 15 degrees. To prevent premature stopping, the stop action cannot be selected within the first 10 steps~\cite{xiao2025uav}. \textit{3) TravelUAV:} is a SOTA method for UAV-VLN. Given multi-view images and high-level instruction with target descriptions and direction, it leverages the multimodal understanding capability of VLM to jointly process visual and textual information for trajectory generation~\cite{wang2024towards}. 

\noindent
\textbf{Metrics.}
For a rigorous evaluation of UAV agents, we employ four widely adopted metrics: Success Rate (SR), Oracle Success Rate (OSR), Navigation Error (NE), and Success-weighted Path Length (SPL), which together quantify task completion, trajectory efficiency, and navigation precision.

\begin{table*}
\caption{Comparison with baseline methods.}
\label{comparison_studies}
\centering
\begin{tabular}{ccccccccccccc} 
  \toprule
  \multirow{2}{*}{\textbf{Methods}} & \multicolumn{4}{c}{Test} & \multicolumn{4}{c}{Test UO}  &  \multicolumn{4}{c}{Test US}\\
  \cmidrule(lr){2-5}   \cmidrule(lr){6-9}   \cmidrule(lr){10-13} 
  &SR \textuparrow & OSR \textuparrow &NE \textdownarrow & SPL \textuparrow &SR \textuparrow & OSR \textuparrow &NE \textdownarrow & SPL \textuparrow &SR \textuparrow & OSR \textuparrow &NE \textdownarrow & SPL \textuparrow  \\
  \midrule
Random&  0.66 & 0.66 & 46.34 & 0.07 & 0.00 & 1.22 & 60.66 & 0.00 & 0.00 & 1.73 & 61.45 & 0.00 \\
Fixed& 0.00 & 3.29  & 46.30 & 0.00 & 0.00 & 2.44 & 60.51 & 0.00 & 0.00 & 0.00 & 59.54 & 0.00  \\
Qwen2.5-VL-7B & 0.00 & 3.29  & 42.31 & 0.00 & 0.00 & 3.05 & 56.53 & 0.00 & 0.58 & 2.31 & 55.36 & 0.04 \\
TravelUAV& 25.00   & 30.26 & 34.03  & 17.89& 20.12 & 24.39 & 37.30 & 16.08 & 16.76 & 24.28 & 38.75 & 12.15 \\
  \midrule
3DG-VLN & \textbf{38.82} & \textbf{55.92} & \textbf{24.41} & \textbf{27.70} & \textbf{28.05} & \textbf{39.63} & \textbf{35.07} & \textbf{21.92} & \textbf{21.39} & \textbf{35.84} & \textbf{37.73} & \textbf{13.43} \\
    \midrule
\end{tabular}
\end{table*}

\subsection{Implementation Details}
We adopt Qwen2.5-VL-7B as the base model for 3DG-VLN to balance navigation capability and computational cost. All training experiments are conducted using the PyTorch framework on an NVIDIA Tesla A100 GPU with 40GB of memory. We set the camera field of view to \(\mathrm{FOV}=\pi/2\), the image resolution to \(W=1024\) and \(H=1024\), the angular thresholds to \(\gamma_{\mathrm{front}}=15^\circ\) and \(\phi=\mathrm{FOV}/4\), and the length of waypoint sequence to \(K=5\). Both 3DG-VLN and TravelUAV are fine-tuned for 2 epochs. For 3DG-VLN, we use a batch size of 2, 8 gradient accumulation steps, a learning rate of \(10^{-4}\) and a warmup ratio of 0.1. To enable a fair comparison of navigation models, both TravelUAV and 3DG-VLN adopt the stopping strategy proposed by TravelUAV~\cite{wang2024towards}. This strategy uses an open-set object detector~\cite{liu2023grounding} to locate the target and retrieves the corresponding depth values within the detected bounding box to determine whether to stop.

\subsection{Comparisons with SOTA Methods} 
\label{comparisons}
We quantitatively compare 3DG-VLN with baseline methods on three distinct test sets, as summarized in Table~\ref{comparison_studies}. Random and Fixed obtain nearly zero SR and OSR across all test sets, indicating that target-visible navigation is still highly challenging due to the vast and unstructured 3D motion space. Although Qwen2.5-VL-7B exhibits limited oracle success in some cases, it still fails to achieve effective target reaching without task-specific waypoint supervision.

Compared with the current SOTA UAV-VLN method, TravelUAV, 3DG-VLN consistently achieves better performance across all test sets and metrics. On the Test set, it improves SR from 25.00\% to 38.82\%, OSR from 30.26\% to 55.92\%, and SPL from 17.89\% to 27.70\%, while reducing NE from 34.03~m to 24.41~m. The improvements on Test UO and Test US further demonstrate its robustness to unseen objects and unseen scenes. These results demonstrate that 3DG-VLN is not only more successful in reaching visible targets, but also more effective in approaching target regions. The improvements mainly stem from its high-resolution egocentric perception, which preserves fine-grained visual evidence for target grounding, and dynamic 3D directional guidance, which provides accurate spatial priors for waypoint-level navigation in the \textit{see-and-reach} stage.

\begin{figure*}
\centering
\includegraphics[width=7in]{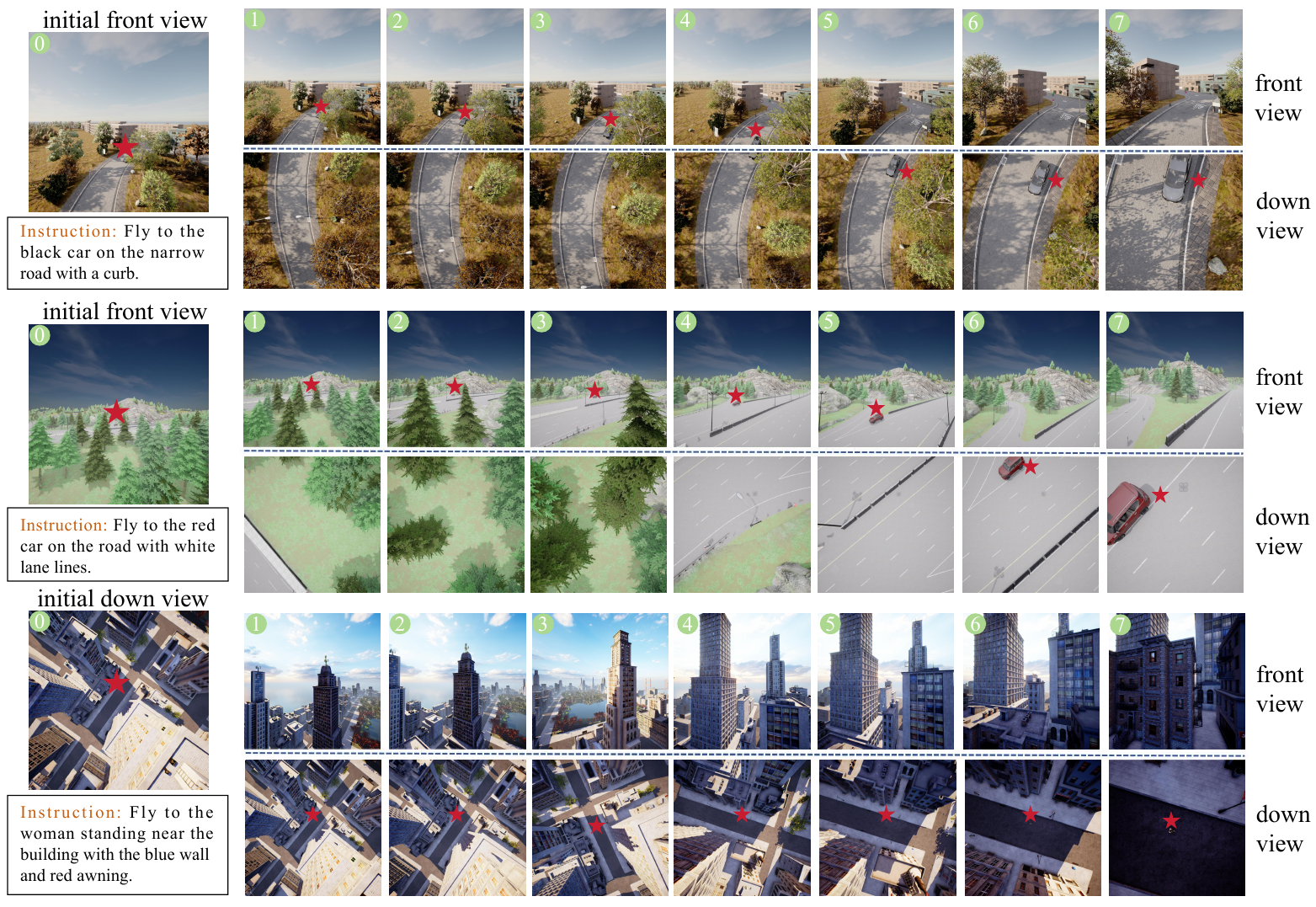}
\caption{Visualization of 3DG-VLN navigation in high-fidelity simulation environments.}
\label{vision}
\end{figure*}

Fig.~\ref{vision} illustrates representative visualized trajectories. Across these examples, 3DG-VLN demonstrates robust navigation and planning capabilities, autonomously generating feasible paths, avoiding obstacles, and efficiently reaching the vicinity of the target.

\begin{figure*}
\centering
\includegraphics[width=7in]{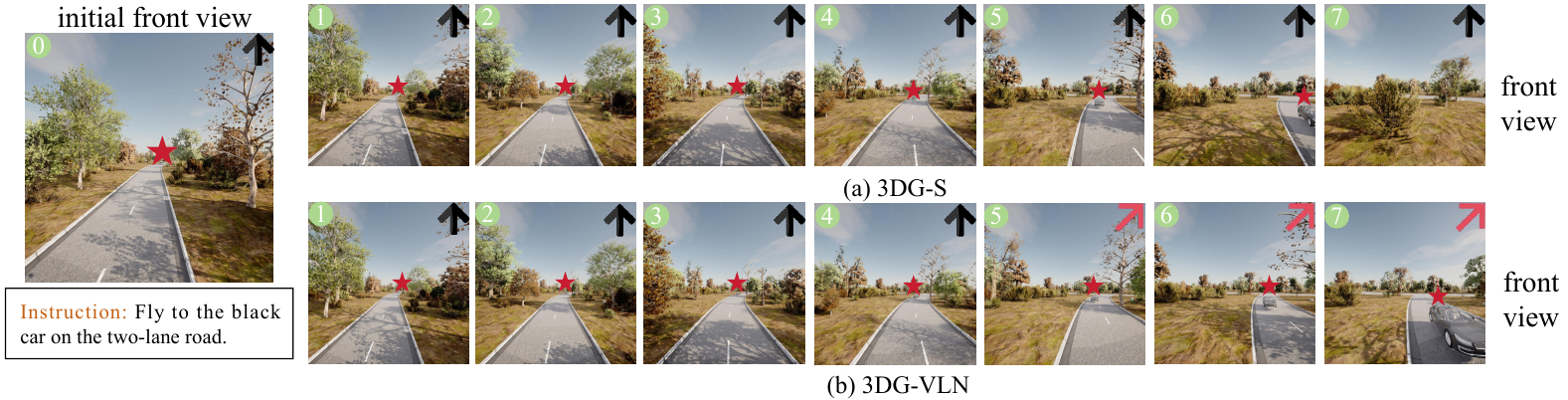}
\caption{Comparison between 3DG-VLN and 3DG-S under the same navigation instruction.
Each sub-image shows the front-view observation at a specific timestep, and the arrow in the upper-right corner indicates the current direction cue, where $\uparrow$ denotes \textit{front} and $\nearrow$ denotes \textit{front-right}.}
\label{online_direction}
\end{figure*}

\begin{table*}
\caption{Ablation study of different strategies. LR:  The model is fine-tuned with low-resolution images resized to \(256 \times 256\), a commonly adopted input resolution in long-range UAV-VLN settings; 3DG-W: Without 3D directional guidance; 3DG-S: Navigating with orientation from the initial instruction only.}
\label{ablation_studies}
\centering
\begin{tabular}{ccccccccccccc} 
  \toprule
  \multirow{2}{*}{\textbf{Strategies}} & \multicolumn{4}{c}{Test} & \multicolumn{4}{c}{Test UO}  &  \multicolumn{4}{c}{Test US}\\
  \cmidrule(lr){2-5}   \cmidrule(lr){6-9}   \cmidrule(lr){10-13} 
  &SR \textuparrow & OSR \textuparrow &NE \textdownarrow & SPL \textuparrow &SR \textuparrow & OSR \textuparrow &NE \textdownarrow & SPL \textuparrow &SR \textuparrow & OSR \textuparrow &NE \textdownarrow & SPL \textuparrow  \\
  \midrule
LR &  23.03 & 36.84 & 35.03 & 17.05 & 18.29 & 23.78 & 38.90 & 13.59 & 17.34 & 29.48 & 39.48 & 10.56 \\
3DG-W& 25.00 & 44.08  & 29.86 & 17.31 & 22.56 & 36.59 & \textbf{33.08} & 17.53 & 17.34 & 31.79 & 43.23 & 10.24  \\
3DG-S& 34.87 & 49.34 & 27.53 & 24.61& 23.17 & 32.93 & 34.08 & 17.09 & 16.76 & 30.06 & 40.52 & 9.19 \\
  \midrule
3DG-VLN & \textbf{38.82} & \textbf{55.92} & \textbf{24.41} & \textbf{27.70} & \textbf{28.05} & \textbf{39.63} & 35.07 & \textbf{21.92} & \textbf{21.39} & \textbf{35.84} & \textbf{37.73} & \textbf{13.43}  \\
    \midrule
\end{tabular}
\end{table*}

\begin{table*}
\caption{Ablation study on outdated and recomputed direction cues. 
TravelUAV-O and 3DG-VLN-O denote the original direction cue computed at the beginning of the long-horizon UAV-VLN task, while TravelUAV and 3DG-S denote the direction cue recomputed at the UAV-VLN-FOV stage using the same direction-computation rule.}
\label{ablation_studies_init_direction}
\centering
\begin{tabular}{ccccccccccccc} 
  \toprule
  \multirow{2}{*}{\textbf{Strategies}} & \multicolumn{4}{c}{Test} & \multicolumn{4}{c}{Test UO}  &  \multicolumn{4}{c}{Test US}\\
  \cmidrule(lr){2-5}   \cmidrule(lr){6-9}   \cmidrule(lr){10-13} 
  &SR \textuparrow & OSR \textuparrow &NE \textdownarrow & SPL \textuparrow &SR \textuparrow & OSR \textuparrow &NE \textdownarrow & SPL \textuparrow &SR \textuparrow & OSR \textuparrow &NE \textdownarrow & SPL \textuparrow  \\
  \midrule
TravelUAV-O &  24.34 & 26.97 & 34.81 & 16.58 & \textbf{20.12} & 22.56 & 37.96 & 15.60 & 16.18 & 23.12 & 39.59 & 11.85 \\
TravelUAV& \textbf{25.00} & \textbf{30.26}  & \textbf{34.03} & \textbf{17.89} & \textbf{20.12} & \textbf{24.39} & \textbf{37.30} & \textbf{16.08} & \textbf{16.76} & \textbf{24.28} & \textbf{38.75} & \textbf{12.15} \\
\midrule
3DG-VLN-O &  26.32 & 40.79 & 35.90 & 19.29 & 17.07 & 26.22 & 36.92 & 12.49 & 14.45 & 26.59 & 43.19 & 8.94 \\
3DG-S& \textbf{34.87} & \textbf{49.34} & \textbf{27.53} & \textbf{24.61}& \textbf{23.17} & \textbf{32.93} & \textbf{34.08} & \textbf{17.09} & \textbf{16.76} & \textbf{30.06} & \textbf{40.52} & \textbf{9.19} \\
\midrule
\end{tabular}
\end{table*}

\subsection{Ablation Study} \label{ablation}
\textbf{Effect of Resolution:} As shown in Table~\ref{ablation_studies}, compared with the LR method, 3DG-VLN achieves significant performance improvements on all metrics across three different test sets. In particular, for the SR, the improvement margins of 3DG-VLN over LR on the three datasets reach 15.79\%, 9.76\% and 4.05\%, respectively; while for the OSR, the margins are 19.08\%, 15.85\% and 6.36\%, respectively. This indicates that the model fine-tuned with high-resolution images can provide more informative visual evidence for target grounding and waypoint prediction. Essentially, high-resolution images can provide fine-grained visual cues regarding target morphology and local spatial structures, thereby supporting more accurate language grounding and 3D motion prediction. This not only confirms the important impact of high-resolution images on the UAV-VLN-FOV task, but also fully demonstrates the necessity of constructing high-resolution datasets for this task.

\textbf{Effect of Directional Guidance:}
Table~\ref{ablation_studies} evaluates the contribution of 3D directional guidance. Compared with 3DG-W, which removes the direction cue, 3DG-VLN improves SR, OSR, and SPL across all test sets, while reducing NE on Test and Test US. Specifically, on the Test set, SR, OSR, and SPL increase from 25.00\%, 44.08\%, and 17.31\% to 38.82\%, 55.92\%, and 27.70\%, respectively. Similar gains are observed on Test UO and Test US, although NE slightly increases on Test UO. This may be because the direction-guided policy in 3DG-VLN is more actively target-seeking based on current visual evidence. Under unseen-object conditions, target grounding can be more uncertain. Therefore, noisy or ambiguous direction cues may occasionally guide the UAV toward an incorrect or visually similar region, leading to larger terminal deviations in a few failure cases. Nevertheless, the consistent gains in SR, OSR, and SPL indicate that 3D directional guidance provides an effective spatial prior for target reaching in the \textit{see-and-reach} stage.

\textbf{Effect of Direction Updating:}
We first examine whether the direction cue should be updated when entering the \textit{see-and-reach} stage. As shown in Table~\ref{ablation_studies_init_direction}, using the direction recomputed at the UAV-VLN-FOV stage consistently outperforms using the outdated direction from the beginning of long-range navigation for both TravelUAV and 3DG-VLN. This suggests that the initial direction may no longer reflect the current UAV-target spatial relation after the UAV has moved, and that direction cues should be updated according to the target-visible reaching stage. Although such stage-level recomputation still provides only a static cue, it already offers a more reliable spatial prior than the stale long-range direction.

We further evaluate whether this direction cue should be continuously updated during closed-loop navigation. As shown in Table~\ref{ablation_studies}, 3DG-VLN consistently outperforms 3DG-S across all test sets, except for a slight increase in NE on Test UO. This confirms that online updating of direction cues helps maintain alignment with the target during the flight process, providing clearer visual cues of the target and enhancing waypoint-level motion prediction, which ultimately improves the UAV’s precision and success in reaching the target within its field of view. The visualization in Fig.~\ref{online_direction} further illustrates this effect. 3DG-VLN dynamically updates the direction cue from \textit{front} to \textit{front-right} as the target shifts in the egocentric view, enabling the UAV to adaptively adjust its motion toward the target, whereas 3DG-S maintains a static direction and gradually loses target alignment.

\begin{figure*}
\centering
\includegraphics[width=7in]{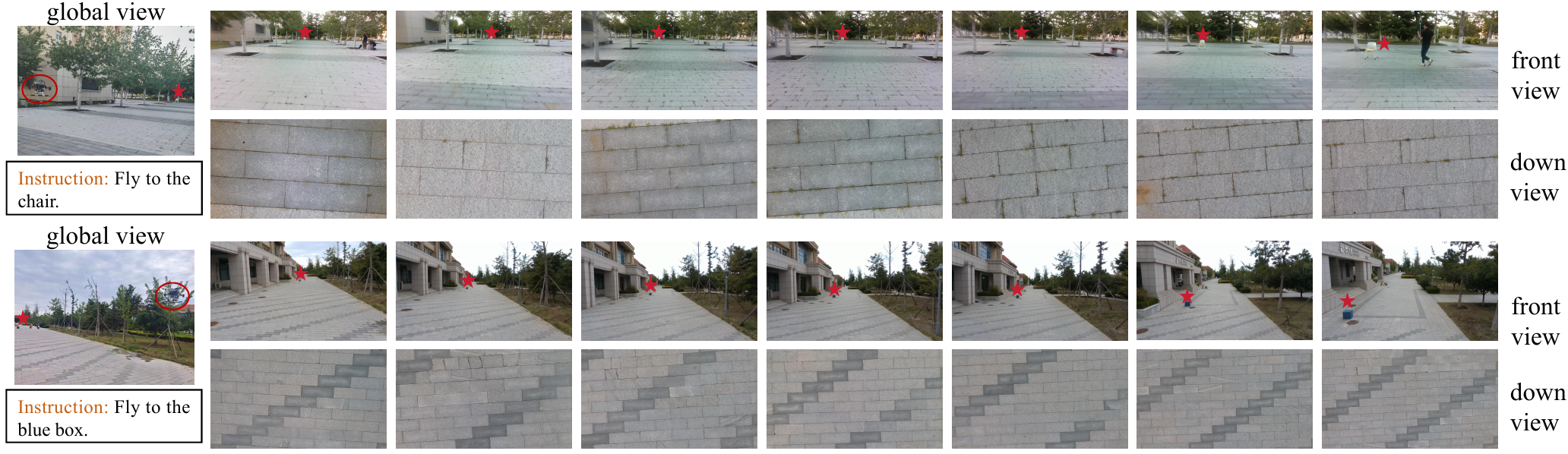}
\caption{Visualization of 3DG-VLN navigation in the real world.}
\label{real_visual_1}
\end{figure*}

\subsection{Real-World Experiments}\label{real_world}
We further conduct real-world experiments to evaluate the practical applicability of 3DG-VLN in precise \textit{see-and-reach} navigation. As shown in Fig.~\ref{real_visual_1}, given the navigation instructions, the UAV predicts continuous 3D waypoints from real-time egocentric observations and progressively navigates toward the targets. Despite real-world challenges such as illumination changes and background variations, 3DG-VLN successfully guides the UAV into a close spatial vicinity of the specified target, rather than merely approaching the target area. This result demonstrates that 3DG-VLN can accurately transform linguistic instructions and visual observations into precise 3D maneuvers for high-precision target reaching, aligning with the stringent 10-meter success criterion emphasized in the UAV-VLN-FOV setting. It also provides preliminary evidence of 3DG-VLN's generalization from simulation to real-world scenarios.

\section{Conclusion and Future Work}
This paper introduces \textbf{UAV-VLN-FOV}, a novel UAV-VLN task that explicitly focuses on the \textit{see-and-reach} stage, where the target is initially visible within the UAV’s field of view. To overcome limitations of existing methods in fine-grained visual grounding and spatial direction alignment, we propose \textbf{3DG-VLN}, a vision-language waypoint prediction framework that integrates high-resolution dual-view observations for fine-grained target grounding and motion prediction, while dynamically updating direction cues during flight to ensure continuous spatial alignment. To support UAV-VLN-FOV, we construct a high-resolution navigation dataset, comprising trajectories annotated with target-oriented high-level instructions and continuous 3D waypoints. Experimental results show that 3DG-VLN significantly outperforms existing baseline methods on this task, achieving SOTA performance.

Despite outperforming the baselines, 3DG-VLN still relies on an external open-set object detector for online direction updating, making target-relative direction estimation sensitive to detection quality. Future work will explore an integrated framework that jointly models visual grounding, spatial alignment, and motion planning to improve precise target-reaching performance.

\bibliographystyle{IEEEtran}
\bibliography{ref}

@article{lin2025openvln,
  title={OpenVLN: Open-world Aerial Vision-Language Navigation},
  author={Lin, Peican and Sun, Gan and Liu, Chenxi and Li, Fazeng and Ren, Weihong and Cong, Yang},
  journal={arXiv preprint arXiv:2511.06182},
  year={2025}
}

@inproceedings{li2025skyvln,
  title={SkyVLN: Vision-and-Language Navigation and NMPC Control for UAVs in Urban Environments},
  author={Li, Tianshun and Huai, Tianyi and Li, Zhen and Gao, Yichun and Li, Haoang and Zheng, Xinhu},
  booktitle={2025 IEEE/RSJ International Conference on Intelligent Robots and Systems (IROS)},
  pages={17199--17206},
  year={2025},
  organization={IEEE}
}

@inproceedings{liu2023aerialvln,
  title={Aerialvln: Vision-and-language navigation for uavs},
  author={Liu, Shubo and Zhang, Hongsheng and Qi, Yuankai and Wang, Peng and Zhang, Yanning and Wu, Qi},
  booktitle={Proceedings of the IEEE/CVF International Conference on Computer Vision},
  pages={15384--15394},
  year={2023}
}

@article{ge2025airsim360,
  title={Airsim360: A panoramic simulation platform within drone view},
  author={Ge, Xian and Pan, Yuling and Zhang, Yuhang and Li, Xiang and Zhang, Weijun and Zhang, Dizhe and Wan, Zhaoliang and Lin, Xin and Zhang, Xiangkai and Liang, Juntao and others},
  journal={arXiv preprint arXiv:2512.02009},
  year={2025}
}

@inproceedings{fan2023aerial,
  title={Aerial vision-and-dialog navigation},
  author={Fan, Yue and Chen, Winson and Jiang, Tongzhou and Zhou, Chun and Zhang, Yi and Wang, Xin},
  booktitle={Findings of the Association for Computational Linguistics: ACL 2023},
  pages={3043--3061},
  year={2023}
}

@article{lee2024citynav,
  title={Citynav: Language-goal aerial navigation dataset with geographic information},
  author={Lee, Jungdae and Miyanishi, Taiki and Kurita, Shuhei and Sakamoto, Koya and Azuma, Daichi and Matsuo, Yutaka and Inoue, Nakamasa},
  journal={arXiv preprint arXiv:2406.14240},
  year={2024}
}

@article{hu2022sensaturban,
  title={Sensaturban: Learning semantics from urban-scale photogrammetric point clouds},
  author={Hu, Qingyong and Yang, Bo and Khalid, Sheikh and Xiao, Wen and Trigoni, Niki and Markham, Andrew},
  journal={International Journal of Computer Vision},
  volume={130},
  number={2},
  pages={316--343},
  year={2022},
  publisher={Springer}
}

@article{liu2024navagent,
  title={Navagent: Multi-scale urban street view fusion for uav embodied vision-and-language navigation},
  author={Liu, Youzhi and Yao, Fanglong and Yue, Yuanchang and Xu, Guangluan and Sun, Xian and Fu, Kun},
  journal={arXiv preprint arXiv:2411.08579},
  year={2024}
}

@inproceedings{xiao2025uav,
  title={Uav-on: A benchmark for open-world object goal navigation with aerial agents},
  author={Xiao, Jianqiang and Sun, Yuexuan and Shao, Yixin and Gan, Boxi and Liu, Rongqiang and Wu, Yanjin and Guan, Weili and Deng, Xiang},
  booktitle={Proceedings of the 33rd ACM International Conference on Multimedia},
  pages={13023--13029},
  year={2025}
}

@article{xu2025geonav,
  title={Geonav: Empowering mllms with explicit geospatial reasoning abilities for language-goal aerial navigation},
  author={Xu, Haotian and Hu, Yue and Gao, Chen and Zhu, Zhengqiu and Zhao, Yong and Li, Yong and Yin, Quanjun},
  journal={arXiv preprint arXiv:2504.09587},
  year={2025}
}

@article{zhang2025grounded,
  title={Grounded Vision-Language Navigation for UAVs with Open-Vocabulary Goal Understanding},
  author={Zhang, Yuhang and Yu, Haosheng and Xiao, Jiaping and Feroskhan, Mir},
  journal={arXiv preprint arXiv:2506.10756},
  year={2025}
}

@article{cai2025flightgpt,
  title={FlightGPT: Towards Generalizable and Interpretable UAV Vision-and-Language Navigation with Vision-Language Models},
  author={Cai, Hengxing and Dong, Jinhan and Tan, Jingjun and Deng, Jingcheng and Li, Sihang and Gao, Zhifeng and Wang, Haidong and Su, Zicheng and Sumalee, Agachai and Zhong, Renxin},
  journal={arXiv preprint arXiv:2505.12835},
  year={2025}
}

@article{liu2025deepseek,
title={Deepseek-v3. 2: Pushing the frontier of open large language models},
author={Liu, Aixin and Mei, Aoxue and Lin, Bangcai and Xue, Bing and Wang, Bingxuan and Xu, Bingzheng and Wu, Bochao and Zhang, Bowei and Lin, Chaofan and Dong, Chen and others},
journal={arXiv preprint arXiv:2512.02556},
year={2025}
}

@misc{bai2025qwen2,
      title={Qwen2.5-VL Technical Report}, 
      author={Shuai Bai and Keqin Chen and Xuejing Liu and Jialin Wang and Wenbin Ge and Sibo Song and Kai Dang and Peng Wang and Shijie Wang and Jun Tang and Humen Zhong and Yuanzhi Zhu and Mingkun Yang and Zhaohai Li and Jianqiang Wan and Pengfei Wang and Wei Ding and Zheren Fu and Yiheng Xu and Jiabo Ye and Xi Zhang and Tianbao Xie and Zesen Cheng and Hang Zhang and Zhibo Yang and Haiyang Xu and Junyang Lin},
      year={2025},
      eprint={2502.13923},
      archivePrefix={arXiv},
      primaryClass={cs.CV}
}

@inproceedings{
hu2022lora,
title={Lo{RA}: Low-Rank Adaptation of Large Language Models},
author={Edward J Hu and Yelong Shen and Phillip Wallis and Zeyuan Allen-Zhu and Yuanzhi Li and Shean Wang and Lu Wang and Weizhu Chen},
booktitle={International Conference on Learning Representations},
year={2022}
}

@inproceedings{wang2024towards,
  title={Towards realistic uav vision-language navigation: Platform, benchmark, and methodology},
  author={Wang, Xiangyu and Yang, Donglin and Kwan, Hohin and Chen, Jinyu and Li, Hongsheng and Liao, Yue and Liu, Si and others},
  booktitle={International Conference on Learning Representations},
  pages={7292--7310},
  year={2025}
}

@inproceedings{liu2023grounding,
  title={Grounding dino: Marrying dino with grounded pre-training for open-set object detection},
  author={Liu, Shilong and Zeng, Zhaoyang and Ren, Tianhe and Li, Feng and Zhang, Hao and Yang, Jie and Jiang, Qing and Li, Chunyuan and Yang, Jianwei and Su, Hang and others},
  booktitle={European conference on computer vision},
  pages={38--55},
  year={2024},
  organization={Springer}
}

@article{zhang2026you, title={What You See Is What You Reach: Towards Spatial Navigation with High-Level Human Instructions}, volume={40}, DOI={10.1609/aaai.v40i15.38258}, abstractNote={Embodied navigation is a fundamental capability that enables embodied agents to effectively interact with the physical world in various complex environments. However, a significant gap remains between current embodied navigation tasks and real-world requirements, as existing methods often struggle to integrate high-level human instructions with spatial understanding. To address this gap, we propose a new task of embodied navigation called spatial navigation, which encompasses two key components: spatial object navigation (SpON) for object-specific guidance and spatial area navigation (SpAN) for navigating to designated areas. Specifically, SpON guides agents to specific objects by leveraging spatial relationships and contextual understanding, while SpAN focuses on navigating to defined areas within complex environments. Together, these components significantly enhance agents’ navigation capabilities, enabling more effective interactions in real-world scenarios. To support this task, we have generated a spatial navigation dataset consisting of 10K trajectories within the simulator. This dataset includes high-level human instructions, detailed observations, and corresponding navigation actions, providing a comprehensive resource to enhance agent training and performance. Building on the spatial navigation dataset, we introduce SpNav, a hierarchical navigation framework. Specifically, SpNav employs vision-language model (VLM) to interpret high-level human instructions and accurately identify goal objects or areas within the observation range, achieving precise point-to-point navigation using a map and enhancing the agent’s ability to oper-
ate effectively in complex environments by bridging the gap between perception and action. Extensive experiments show that SpNav achieves state-of-the-art (SOTA) performance in spatial navigation tasks across both simulated and real-world environments, validating the effectiveness of our method.}, number={15}, journal={Proceedings of the AAAI Conference on Artificial Intelligence}, author={Zhang, Lingfeng and Fu, Haoxiang and Hao, Xiaoshuai and Zhang, Shuyi and Zhang, Qiang and Liu, Rui and Chen, Long and Ding, Wenbo}, year={2026}, month={Mar.}, pages={12627–12635} }

@inproceedings{zhang2025citynavagent,
    title = "{C}ity{N}av{A}gent: Aerial Vision-and-Language Navigation with Hierarchical Semantic Planning and Global Memory",
    author = "Zhang, Weichen  and
      Gao, Chen  and
      Yu, Shiquan  and
      Peng, Ruiying  and
      Zhao, Baining  and
      Zhang, Qian  and
      Cui, Jinqiang  and
      Chen, Xinlei  and
      Li, Yong",
    editor = "Che, Wanxiang  and
      Nabende, Joyce  and
      Shutova, Ekaterina  and
      Pilehvar, Mohammad Taher",
    booktitle = "Proceedings of the 63rd Annual Meeting of the Association for Computational Linguistics (Volume 1: Long Papers)",
    month = jul,
    year = "2025",
    address = "Vienna, Austria",
    publisher = "Association for Computational Linguistics",
    pages = "31292--31309",
    ISBN = "979-8-89176-251-0"
}

@inproceedings{wang2026expand,
  title={Expand Your SCOPE: Semantic Cognition over Potential-Based Exploration for Embodied Visual Navigation},
  author={Wang, Ningnan and Chen, Weihuang and Chen, Liming and Ji, Haoxuan and Guo, Zhongyu and Zhang, Xuchong and Sun, Hongbin},
  booktitle={Proceedings of the AAAI Conference on Artificial Intelligence},
  volume={40},
  number={22},
  pages={18620--18628},
  year={2026}
}

@article{he2026fine,
  title={Fine-grained alignment supervision matters in vision-and-language navigation},
  author={He, Keji and Huang, Yan and Jing, Ya and Wu, Qi and Wang, Liang},
  journal={IEEE Transactions on Pattern Analysis and Machine Intelligence},
  year={2026},
  publisher={IEEE}
}

@article{ning2026lookasidevln,
  title={LookasideVLN: direction-aware aerial vision-and-language navigation},
  author={Ning, Yuwei and Zhao, Ganlong and Qin, Yipeng and Liu, Si and Liu, Yang and Lin, Liang and Li, Guanbin},
  journal={arXiv preprint arXiv:2604.17190},
  year={2026}
}

@inproceedings{liu2025vln,
  title={VLN-ChEnv: Vision-language Navigation in Changeable Environments},
  author={Liu, Shubo and Zhang, Hongsheng and Qiao, Qian and Wu, Qi and Wang, Peng},
  booktitle={Proceedings of the 33rd ACM International Conference on Multimedia},
  pages={3798--3807},
  year={2025}
}

@inproceedings{zhang2025multimodal,
  title={Multimodal inverse attention network with intrinsic discriminant feature exploitation for fake news detection},
  author={Zhang, Tianlin and Yu, En and Shao, Yi and Sun, Jiande},
  booktitle={Proceedings of the Thirty-Fourth International Joint Conference on Artificial Intelligence},
  pages={7940--7948},
  year={2025}
}

@inproceedings{yu2026generalized,
  title={Generalized incremental learning under concept drift across evolving data streams},
  author={Yu, En and Lu, Jie and Zhang, Guangquan},
  booktitle={Proceedings of the ACM Web Conference 2026},
  pages={3905--3916},
  year={2026}
}

@article{xu2026vgas,
  title={VGAS: Value-Guided Action-Chunk Selection for Few-Shot Vision-Language-Action Adaptation},
  author={Xu, Changhua and Yu, En and Xuan, Junyu and Lu, Jie},
  journal={arXiv preprint arXiv:2602.07399},
  year={2026}
}

@inproceedings{ding2026history,
  title={History-enhanced two-stage transformer for aerial vision-and-language navigation},
  author={Ding, Xichen and Gao, Jianzhe and Pan, Cong and Wang, Wenguan and Qin, Jie},
  booktitle={Proceedings of the AAAI Conference on Artificial Intelligence},
  volume={40},
  number={22},
  pages={18225--18233},
  year={2026}
}

@article{xu2026aerialvla,
  title={AerialVLA: A Vision-Language-Action Model for UAV Navigation via Minimalist End-to-End Control},
  author={Xu, Peng and Deng, Zhengnan and Deng, Jiayan and Gu, Zonghua and Wan, Shaohua},
  journal={arXiv preprint arXiv:2603.14363},
  year={2026}
}

@article{zheng2026onfly,
  title={OnFly: Onboard Zero-Shot Aerial Vision-Language Navigation toward Safety and Efficiency},
  author={Zheng, Guiyong and Ban, Yueting and Zhang, Mingjie and Zheng, Juepeng and Zhou, Boyu},
  journal={arXiv preprint arXiv:2603.10682},
  year={2026}
}

@inproceedings{chen2026aerialvla,
  title={AerialVLA: A Vision-Language-Action Model for Aerial Navigation with Online Dialogue},
  author={Chen, Jinyu and Li, Hongyu and Tang, Zongheng and Li, Xiaoduo and Wu, Wenjun and Liu, Si},
  booktitle={Proceedings of the AAAI Conference on Artificial Intelligence},
  volume={40},
  number={22},
  pages={18161--18169},
  year={2026}
}

@article{cai2026airnav,
  title={AirNav: A Large-Scale Real-World UAV Vision-and-Language Navigation Dataset with Natural and Diverse Instructions},
  author={Cai, Hengxing and Rao, Yijie and Huang, Ligang and Zhong, Zanyang and Dong, Jinhan and Tan, Jingjun and Lu, Wenhao and Zhong, Renxin},
  journal={arXiv preprint arXiv:2601.03707},
  year={2026}
}

@article{wang2026dynam3d,
  title={Dynam3d: Dynamic layered 3d tokens empower vlm for vision-and-language navigation},
  author={Wang, Zihan and Lee, Seungjun and Lee, Gim Hee},
  journal={Advances in Neural Information Processing Systems},
  volume={38},
  pages={153522--153544},
  year={2026}
}

@article{zhang2025flexvln,
  title={Flexvln: Flexible adaptation for diverse vision-and-language navigation tasks},
  author={Zhang, Siqi and Qiao, Yanyuan and Wang, Qunbo and Guo, Longteng and Wei, Zhihua and Liu, Jing},
  journal={IEEE Transactions on Multimedia},
  volume={27},
  pages={6307--6318},
  year={2025},
  publisher={IEEE}
}

@inproceedings{zhang2025cosmo,
  title={Cosmo: Combination of selective memorization for low-cost vision-and-language navigation},
  author={Zhang, Siqi and Qiao, Yanyuan and Wang, Qunbo and Yan, Zike and Wu, Qi and Wei, Zhihua and Liu, Jing},
  booktitle={Proceedings of the IEEE/CVF International Conference on Computer Vision},
  pages={5511--5522},
  year={2025}
}

@article{wu2025vla,
  title={VLA-AN: An Efficient and Onboard Vision-Language-Action Framework for Aerial Navigation in Complex Environments},
  author={Wu, Yuze and Zhu, Mo and Li, Xingxing and Du, Yuheng and Fan, Yuxin and Li, Wenjun and Han, Zhichao and Zhou, Xin and Gao, Fei},
  journal={arXiv preprint arXiv:2512.15258},
  year={2025}
}

@article{jiang2026spatialfly,
  title={SpatialFly: Geometry-Guided Representation Alignment for UAV Vision-and-Language Navigation in Urban Environments},
  author={Jiang, Wen and Huang, Kangyao and Wang, Li and Xu, Wang and Fan, Wei and Liu, Jinyuan and Liu, Shaoyu and Liang, Hanfang and Duan, Hongwei and Xu, Bin and others},
  journal={arXiv preprint arXiv:2603.21046},
  year={2026}
}

@ARTICLE{xue2025acds,
  author={Xue, Fanfu and Sun, Jiande and Xue, Yaqi and Wu, Qiang and Zhu, Lei and Chang, Xiaojun and Cheung, Sen-Ching},
  journal={IEEE Transactions on Image Processing}, 
  title={Attention Guidance by Cross-Domain Supervision Signals for Scene Text Recognition}, 
  year={2025},
  volume={34},
  number={},
  pages={717-728},
  doi={10.1109/TIP.2024.3523799}}

@article{yu2026enhancing,
  title={Enhancing outdoor vision: Binocular desnowing with dual-stream temporal transformer},
  author={Yu, En and Lu, Jie and Zhang, Kaihao and Zhang, Guangquan},
  journal={Pattern Recognition},
  volume={170},
  pages={112075},
  year={2026},
  publisher={Elsevier}
}

@article{yu2025mossvln,
  title={MossVLN: Memory-Observation Synergistic System for Continuous Vision-Language Navigation},
  author={Yu, Ting and Wu, Yifei and Cui, Qiongjie and Huang, Qingming and Yu, Jun},
  journal={IEEE Transactions on Multimedia},
  volume={27},
  pages={6690-6704},
  year={2025},
  publisher={IEEE}
}

@article{wu2024vision,
  title={Vision-and-language navigation via latent semantic alignment learning},
  author={Wu, Siying and Fu, Xueyang and Wu, Feng and Zha, Zheng-Jun},
  journal={IEEE Transactions on Multimedia},
  volume={26},
  pages={8406--8418},
  year={2024},
  publisher={IEEE}
}

@article{tan2025source,
  title={Source-free elastic model adaptation for vision-and-language navigation},
  author={Tan, Mingkui and Chen, Peihao and Zhi, Hongyan and Mai, Jiajie and Rosman, Benjamin and Ji, Dongyu and Zeng, Runhao},
  journal={IEEE Transactions on Multimedia},
  volume={27},
  pages={3953-3965},
  year={2025},
  publisher={IEEE}
}

@article{yuan2025seeing,
  title={Seeing with Words: Interpretable Language-Guided Drone Geo-localization via LLM-Enriched Semantic Attribute Alignment},
  author={Yuan, Changsen and Zhou, Yang-Hao and Guo, Cunhan and Han, Danjie and Shi, Ge and Wang, Wenwu},
  journal={IEEE Transactions on Multimedia},
  volume={28},
  pages={2132-2144},
  year={2025},
  publisher={IEEE}
}

@article{YU2026112075,
title = {Enhancing outdoor vision: Binocular desnowing with dual-stream temporal transformer},
journal = {Pattern Recognition},
volume = {170},
pages = {112075},
year = {2026},
issn = {0031-3203},
author = {En Yu and Jie Lu and Kaihao Zhang and Guangquan Zhang}
}

@inproceedings{wu2025aeroduo,
  title={AeroDuo: Aerial Duo for UAV-based Vision and Language Navigation},
  author={Wu, Ruipu and Zhang, Yige and Chen, Jinyu and Huang, Linjiang and Zhang, Shifeng and Zhou, Xu and Wang, Liang and Liu, Si},
  booktitle={Proceedings of the 33rd ACM International Conference on Multimedia},
  pages={2576--2585},
  year={2025}
}

@inproceedings{zhong2026run,
  title={Run, Ruminate, and Regulate: A Dual-process Thinking System for Vision-and-Language Navigation},
  author={Zhong, Yu and Zhang, Zihao and Zhang, Rui and Huang, Lingdong and Gao, Haihan and Wang, Shuo and Li, Da and Han, Ruijian and Guo, Jiaming and Peng, Shaohui and others},
  booktitle={Proceedings of the AAAI Conference on Artificial Intelligence},
  volume={40},
  number={22},
  pages={18845--18854},
  year={2026}
}

@article{wang2025uav,
  title={UAV-Flow Colosseo: A Real-World Benchmark for Flying-on-a-Word UAV Imitation Learning},
  author={Wang, Xiangyu and Yang, Donglin and Liao, Yue and Zheng, Wenhao and Dai, Bin and Li, Hongsheng and Liu, Si and others},
  journal={arXiv preprint arXiv:2505.15725},
  year={2025}
}

\vfill

\end{document}